\documentclass[a4paper,fleqn]{cas-dc}
\usepackage[numbers]{natbib}

\def\tsc#1{\csdef{#1}{\textsc{\lowercase{#1}}\xspace}}
\tsc{WGM}
\tsc{QE}
\usepackage{caption}
\usepackage{graphicx}
\usepackage{multirow}
\usepackage{hhline}
\usepackage{arydshln}
\usepackage{pifont}
\usepackage{algorithm,algorithmicx,listings}        
\usepackage{enumitem}
\usepackage{xcolor}
\usepackage[noend]{algpseudocode}	
\definecolor{first}{RGB}{114,188,213}
\definecolor{second}{RGB}{170,220,224}
\begin{document}
\let\WriteBookmarks\relax
\def\floatpagepagefraction{1}
\def\textpagefraction{.001}
\let\printorcid\relax 

\shorttitle{}    

\shortauthors{Yiming Ji et al.}

\title[mode = title]{Diffusion as Reasoning: Enhancing Object Navigation via Diffusion Model Conditioned on LLM-based Object-Room Knowledge}  

\author[1]{Yiming Ji}
\ead{jiyiming@alu.hit.edu.cn}
\author[1]{Kaijie Yun}
\author[1]{Yang Liu}
\ead{liuyanghit@hit.edu.cn}
\cormark[1]
\author[1]{Zhengpu Wang}
\author[1,2]{Boyu Ma}
\author[1]{Zongwu Xie}
\author[1]{Hong Liu}

\address[1]{State Key Laboratory of Robotics and Systems, Harbin Institute of Technology, Harbin 150001, China} 
\address[2]{School of Mechanical and Aerospace Engineering, Nanyang Technological University, Singapore 639798, Singapore}
\cortext[1]{Corresponding author}  

\begin{abstract}
The Object Navigation (ObjectNav) task aims to guide an agent to locate target objects in unseen environments using partial observations. Prior approaches have employed location prediction paradigms to achieve long-term goal reasoning, yet these methods often struggle to effectively integrate contextual relation reasoning. Alternatively, map completion-based paradigms predict long-term goals by generating semantic maps of unexplored areas. However, existing methods in this category fail to fully leverage known environmental information, resulting in suboptimal map quality that requires further improvement.
In this work, we propose a novel approach to enhancing the ObjectNav task, by training a diffusion model to learn the statistical distribution patterns of objects in semantic maps, and using the map of the explored regions during navigation as the condition to generate the map of the unknown regions, thereby realizing the long-term goal reasoning of the target object, i.e., diffusion as reasoning (DAR).
Meanwhile, we propose the Room Guidance method, which leverages commonsense knowledge derived from large language models (LLMs) to guide the diffusion model in generating room-aware object distributions. Based on the generated map in the unknown region, the agent sets the predicted location of the target as the goal and moves towards it. Experiments on Gibson and MP3D show the effectiveness of our method.
\end{abstract}




\maketitle

\section{Introduction}
\label{sec:intro}
Embodied navigation, which requires an agent to use visual sensing to actively interact with its environment and perform navigation tasks, has seen rapid development driven by large-scale photorealistic domestic scene datasets \cite{chang2017matterport3d}, \cite{ramakrishnan2021habitat}, \cite{xia2018gibson} and advanced high-performance simulators \cite{deitke2020robothor}, \cite{savva2019habitat}.
As a specific form of embodied navigation, Object Navigation (ObjectNav) \cite{hu2023agent} is regarded as a key technology for enabling robotic intelligence.
In ObjectNav task, the agent is placed in an unknown environment and is required to navigate to a user-specified object category (e.g., sofa) based on visual observations. Since the environment is unseen, when the target object is not visible, the agent needs to infer the potential locations of the target.

Embodied navigation agents show intelligence through semantic understanding and reasoning.
Previous work explored learning object contextual relationships for ObjectNav systems.
Some methods extract object/region/room relationships to enhance visual representation  \cite{du2020learning}, \cite{druon2020visual}.
while other end-to-end approaches directly map vision to actions \cite{chen2023think}, \cite{mayo2021visual}.
However, end-to-end methods implicitly embed semantic prior knowledge and simultaneously learn localization, mapping, and path planning in an implicit manner, which leads to the problems of high computational load and limited generalization in unseen environments.

Alternatively, modular approaches typically address the ObjectNav problem by incorporating three modules: Mapping, Long-term Goal Policy, and Path Planning. Among these, the Long-term Goal Policy module plays a pivotal role. When the target object is absent from the currently constructed local map, a semantic reasoning algorithm is employed to infer an appropriate location as the long-term goal, which subsequently serves as input to the Path Planning module. Notably, the accuracy of the Long-term Goal Policy critically determines both the success rate and efficiency of the ObjectNav system.

The existing approaches for long-term goal inference can be broadly categorized into two classes: \textbf{\textit{location prediction methods}} and \textbf{\textit{map completion methods}}.

Location prediction methods refer to algorithms that directly predict the position of the long-term goal based on the semantic map constructed from explored regions and the target object category. Some methods perform location prediction along the frontiers of the local map, selecting the point closest to the target object as the long-term goal \cite{ramakrishnan2022poni}, or incorporating common-sense knowledge to enhance frontier-based predictions \cite{sun2025enhancing}.
Other methods extend beyond frontier regions, instead predicting the target object’s location globally and designating it as the long-term goal \cite{chaplot2020object} \cite{zhai2023peanut}. Additionally, Xinyao Yu et al. proposed a trajectory prediction approach \cite{yu2024trajectory}, which leverages the agent’s historical observations and the target object as conditions to predict a sequence of long-term goals. This method, in essence, also falls under the category of location prediction.

However, location prediction methods suffer from two key limitations: (1) their reliance on pre-constructed datasets, and (2) their individual learning of target-object relations, which may vary significantly across different room layouts. These limitations are effectively addressed by map completion methods, which adopt a self-supervised learning paradigm to jointly model inter-object relationships (referred to as \textit{contextual object relations}). 
Recent advances in map completion have employed encoder-decoder architectures to model these contextual object relations.

Several approaches employ CNN-based models (e.g., U-Net \cite{ronneberger2015u}) to predict unobserved regions based on single-timestamp top-down maps \cite{liang2021sscnav} \cite{georgakis2021learning}. Furthermore, SGM \cite{zhang2024imagine} adopts a masked modeling strategy \cite{he2022masked} to train a transformer-based model (ViT \cite{dosovitskiy2020image}), which integrates visual observations with  LLM knowledge (e.g., GPT-4 \cite{achiam2023gpt}, ChatGLM \cite{du2021glm}). This framework operates by first patchifying the input map and then predicting missing patch values conditioned on visible patches.

The map completion method predicts long-term goals by generating semantic maps of unknown regions. Undoubtedly, the higher the reliability of the generated maps, the greater the improvement in the performance of the ObjectNav system. However, the quality of the unknown-area maps generated by the aforementioned methods remains suboptimal. On one hand, these methods focus solely on objects themselves, while the absolute spatial positions of objects may vary across different environments. On the other hand, existing methods fail to fully utilize the information from observed regions.

We observe that learning contextual object relations equates to learning the statistical distribution patterns of object placements within indoor rooms.
Generating the object distribution maps in unknown regions equates to sampling from the distributions that best match those in the observed areas.

Inspired by the remarkable success of diffusion models in representing complex distributions \cite{ho2020denoising}, we propose replacing conventional CNN-based and Transformer-based architectures in self-supervised map generation with a diffusion model framework. Unlike existing approaches that focus solely on object-level features, our method incorporates room category information as conditional guidance.
While object locations may exhibit ambiguity across different room layouts, specific room categories (e.g., bedrooms) consistently contain characteristic object classes (e.g., beds, pillows). By leveraging prior knowledge of a region's room category, our model can more effectively constrain the generation process to avoid implausible object distributions, thereby significantly improving the quality of generated maps for unknown regions.

In this paper, we present a modular solution for the ObjectNav problem, introducing a Diffusion-based Approach for long-term goal Reasoning (DAR). Our framework conditions on the agent's environmental memory while leveraging room category distributions as guidance to generate reliable semantic content in unknown areas. The DAR module selects potential target object locations within these generated regions as long-term goals.
For training DAR, we utilize standard ObjectNav datasets (Gibson, Matterport3D, etc.) to obtain ground-truth semantic room maps. These maps undergo random rotations, mirroring, and translations to construct the training set for our diffusion model. Following the self-supervised learning paradigm of DDPM, our DAR model learns to generate plausible, novel room semantic maps - distinct from training samples - through iterative denoising from pure Gaussian noise.

However, unlike image datasets, training scenes for ObjectNav suffer from notable limitations in both quantity and diversity. To enhance generalization capability, previous approaches have incorporated general knowledge from LLMs. Our DAR module adopts this paradigm through object-room relationship modeling.
The denoising model in DAR takes both the denoised result xt from the previous time step and a room guidance map as input. This approach enables the diffusion model to generate specific object clusters according to room types. Since diffusion models have demonstrated the capability to generate pixel values in unknown regions based on partially known areas without requiring retraining, we leverage this mechanism by using the semantic map of known regions from actual navigation as a condition. The pre-trained diffusion model is employed to generate semantic maps for unknown regions, with room type guidance also applied.
We first determine the room type of frontier regions using common-sense knowledge extracted from LLMs, based on the observed object distribution in the established semantic map. For instance, if objects near frontier point $p$ include \textit{oven}, \textit{sink}, and \textit{refrigerator} ordered by proximity, the LLM would identify $p$ as located in a kitchen. This room type then guides the diffusion model to generate object distributions around $p$ that are consistent with a kitchen (e.g., \textit{cup}, \textit{bottle}) rather than incompatible objects (e.g., \textit{bed}, \textit{tv}).

We evaluate our DAR model in simulation environments, including photorealistic 3D platforms: AI2-Thor, Robo-Thor, Gibson, and MP3D. Experimental results demonstrate that our DAR model significantly outperforms previous map completion methods, achieving substantial improvements in map quality, object diversity, and room structure/contour accuracy.
The precise semantic understanding enables DAR to produce more accurate semantic maps, leading to better long-term goal prediction and more efficient object navigation.

In summary, our study makes the following key contributions:
\begin{itemize}
    \item   We propose a modular-based ObjectNav approach, DAR, that learns statistical distributions of rooms and objects, enabling accurate generation of unknown-region object layouts based on explored-area maps.
    \item   Leveraging room-object correlations extracted from LLMs, we infer potential room types at frontier regions of partial semantic maps.
    \item   We design a denoising model augmented with a room-type guidance submodule, ensuring generated object distributions adhere to inferred spatial semantics.
    \item   Extensive evaluations on Gibson and Matterport3D (MP3D) demonstrate that DAR achieves state-of-the-art Success Rate (SR). Notably, it improves SPL (efficiency metric) by 10.6\% on average over prior works across both datasets.
\end{itemize}
\section{Related Work}

\subsection{Object navigation}
ObjectNav aims to navigate an agent to a specific object in a previously unknown indoor scene.
Existing approaches fall into three categories \cite{sun2024survey}: end-to-end, modular, and zero-shot methods. End-to-end methods use reinforcement learning (RL) \cite{zhang2022generative} or imitation learning (IL) \cite{ramrakhya2023pirlnav} to map observations directly to actions. Previous RL-based methods focus on learning object relations \cite{ye2021hierarchical}, visual representations \cite{du2021vtnet}, auxiliary tasks \cite{ye2021auxiliary}, and data augmentation \cite{maksymets2021thda} to improve navigation. However, these approaches face challenges in real-world application as they require simultaneous localization, mapping, and planning, with additional difficulties in transferring from simulation to real-world environments due to perceptual differences \cite{gervet2023navigating}.

Unlike end-to-end methods, modular approaches \cite{ramakrishnan2022poni}, \cite{zhang2024imagine} break down ObjectNav tasks into separate components: mapping \cite{chaplot2020object}, long-term-goal reasoning \cite{sun2025enhancing}, and path planning \cite{sethian1996fast}. The mapping module provides a general representation that separates perception from reasoning and planning, making it more suitable for real-world applications.

Zero-shot methods involve navigating to unseen targets that only appear during testing \cite{zhao2023zero}. Existing approaches \cite{chen2023not}, \cite{gadre2023cows} use Vision Language Models (VLMs) and Large Language Models (LLMs) to provide prior knowledge about these unseen targets. However, relying solely on pre-trained VLMs or LLMs limits reliability, especially when the robot detects limited semantic information.
Our DAR is a modular method that uses a diffusion model to design a novel long-term goal reasoning module. It also incorporates the common-sense knowledge from LLMs, avoiding the reliability issues that come with relying solely on LLMs.

\subsection{Visual semantic reasoning}
In ObjectNav, visual semantic reasoning is meant to provide high-level information about the location of the target object. Considering the two extremes, when the semantic reasoning module is highly inefficient, the navigation system tends to explore aimlessly until it has traversed the entire room. On the other hand, when the semantic reasoning module can accurately and reliably compute the target's position, it can greatly improve the efficiency of the navigation system by avoiding exploration in ineffective regions and directly reaching the target.

SK Ramakrishnan \cite{ramakrishnan2022poni} proposed two complementary potential functions as a semantic reasoning approach to determine long-term goals in frontier areas.
L2M \cite{georgakis2021learning} actively imagines environmental layouts beyond the agent's field of view and uses uncertainty in the semantic classes of unobserved areas to choose long-term goals.
S Zhang \cite{zhang2024imagine} proposed training with an MAE mechanism to leverage both episodic observations and general knowledge for generating semantic maps of unknown areas, enabling long-term goal reasoning. This approach is similar to our DAR method. However, the MAE mechanism involves random masking of the established local map, causing information loss. In contrast, our DAR method fully utilizes the local map to guide a diffusion model for generating high-quality semantic maps.
\subsection{Diffusion models}
Diffusion models are a type of deep generative model that have recently gained attention for tasks like image generation \cite{dhariwal2021diffusion}, image translation \cite{zhao2022egsde}, inpainting \cite{lugmayr2022repaint}, and editing \cite{avrahami2023blended}. They consist of a forward diffusion stage, where Gaussian noise is gradually added to the data, and a reverse diffusion stage, where the model learns to recover the original data step by step.

Denoising Diffusion Probabilistic Models (DDPMs), a key type of diffusion model, demonstrated the ability to generate high-quality images, with benefits like full distribution coverage, a stable training objective, and easy scalability \cite{ho2020denoising}\cite{nichol2021improved}\cite{song2019generative}.
Our DAR model leverages DDPMs to learn the distribution of objects in rooms, enabling the generation of high-quality room maps. We also adapt DDPM's image inpainting capabilities to predict unknown areas based on known conditions, offering a novel approach to the semantic reasoning needed in ObjectNav systems.

\section{Methodology}
\label{sec:method}

\subsection{Task Definition}
\label{sec:3-A}
The ObjectNav task involves an agent navigating to an instance of a given object category in an unknown environment. At the start of each episode, the agent is randomly placed. At each timestep $t$, it receives egocentric RGB-D observations, target object information, and its current pose ${\left( x,y,\theta \right) }$. The agent can perform one of four discrete actions: \textit{move forward}, \textit{turn left}, \textit{turn right}, or \textit{stop}.
The agent autonomously executes a stop action when it determines the task is complete. An episode is considered successful if, within a given number of steps, the agent stops at a position where the distance to the target is below a threshold (e.g., 1m) and the target is visible in its egocentric view.

Our method builds on the modular ObjectNav architecture, which typically constructs an accumulating semantic map (${ m_t\in \mathbb{R} ^{\left( N_o+N_s \right) \times H\times W}}$) during navigation.
The semantic map uses multiple channels to represent different classes: ${N_o}$ for occupancy classes (occupied and free), and ${N_s}$ for object classes, with $H$ and $W$ as the map dimensions. This map accumulates observed object layouts of the environment from timesteps 0 to $t$.

However, since the semantic map is built from partial observations, it represents only a subset (i.e., a local map) of the entire environment. When the target is not visible, the agent must estimate the unobserved surroundings based on learned contextual relationships among objects to infer the target's likely location.

\subsection{Self-supervised Diffusion Learning}
\label{sec:3-B}
\subsubsection{Data Preparation}
\begin{figure}[!h]
    \begin{center}
        \includegraphics[width=\linewidth]{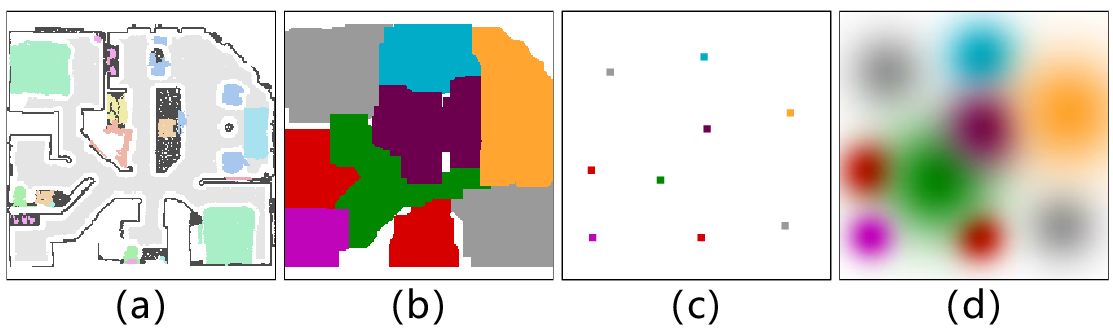}
    \end{center}
    \caption{
        (a) Object semantic map, color encoded; (b) Room semantic segmentation map; (c) Room guidance with points; (d) Room guidance with exponentially decreasing distribution.
        }
    \label{fig:fig01}
\end{figure}

The DAR model is trained in a self-supervised manner. We extract semantic maps for each floor of every scene from the standard ObjectNav dataset. 
To obtain ground-truth room segmentation maps, we note that both the Gibson \cite{xia2018gibson} and Matterport3D \cite{chang2017matterport3d} datasets provide annotated object and room/area categories for each house (e.g., the 3D scene graph \cite{armeni20193d} with Gibson annotations). 
Following the same processing method as LROGNav \cite{sun2025enhancing}, we compute 3D bounding boxes for each room using face IDs and key house parameters (e.g., floors and floor heights), then project the house point cloud into 2D floor maps to derive top-down room segmentation, as shown in Fig. \ref{fig:fig01}(b).

So far, taking Gibson as an example, we have obtained an $H\times W\times 18$ object map $M_o$ (with 18 object categories, Fig. \ref{fig:fig01}(a)) and an  $H\times W\times 19$ room segmentation map $M_r$ (with 19 room categories, Fig. \ref{fig:fig01}(b)).
The DAR model aims to leverage the room segmentation map as guidance in the reverse process, where $M_r$ should provide prior information about the distribution of objects within certain regions.
However, the current $M_r$ contains clear boundaries between different rooms, which, while reasonable in the dataset, may not hold in real-world navigation—agents can hardly infer exact room boundaries from limited observations. Thus, we want $M_r$ to offer object distribution priors without imposing structural constraints on the generated object map.

To address this, we modify $M_r$ as follows: For each room category channel, we first cluster regions of that category, then compute the geometric center ($c_x$, $c_y$) of each cluster. 
Finally, we design a distribution map where pixel values decay exponentially from the center outward, as defined in Eq. \ref{eq:eq01}.
\begin{equation}  
    \label{eq:eq01}
    M_{r}^{i}\left( x,y \right) =\exp \left( -\frac{\sqrt{\left( x-c_x \right) ^2+\left( y-c_y \right) ^2}}{\sigma ^2} \right)
\end{equation}
Here, $M_{r}^{i}$ denotes the i-th channel of the room segmentation map, where $(x, y)$ represents pixel coordinates. For each room cluster, we compute its minimum bounding box and set $\sigma$ (the decay coefficient) as one-fourth of the average side length of the bounding box, and get room guidance as shown in Fig. \ref{fig:fig01}(d).
In Section \ref{sec:4-A}, we evaluate how different approaches for determining the $\sigma$ value affect the quality of the generated maps in unexplored regions.

\subsubsection{Denoising Model Design}

We employ a U-Net-based architecture as our backbone model and investigate three distinct denoising network configurations for the diffusion process, as shown in Fig. \ref{fig:fig02}.

\begin{figure}[!h]
    \begin{center}
        \includegraphics[width=0.8\linewidth]{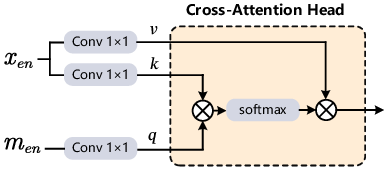}
    \end{center}
    \caption{
        Illustration of the cross-attention mechanism.
        }
    \label{fig:fig03}
\end{figure}

\begin{figure}[!h]
    \begin{center}
        \includegraphics[width=\linewidth]{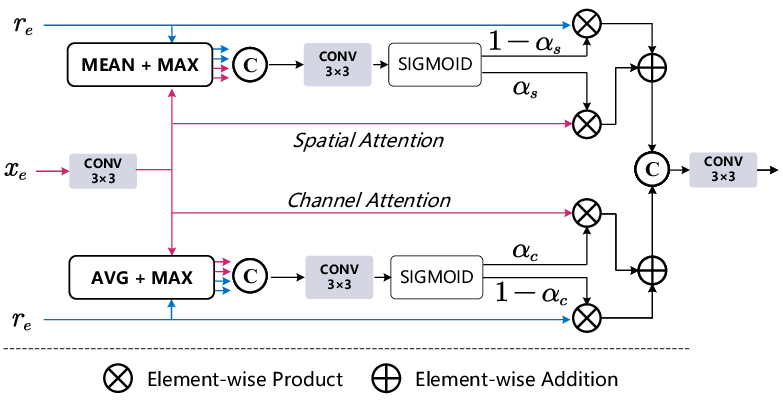}
    \end{center}
    \caption{
        Structure of Semantic Attention module.
        }
    \label{fig:fig04}
\end{figure}

\begin{itemize}
    \item \textbf{\textit{Concatenation}}: The most straightforward approach directly concatenates the input $m_r$ and $x_t$ before feeding them into the U-Net denoising network. While simple to implement, this method significantly increases the channel dimension (to objects\_class\_num + rooms\_class\_num), potentially affecting computational efficiency and model performance.
    \item \textbf{\textit{Cross-Attention}}: We adopt the same architecture as the U-Net encoder blocks to construct the Room Guidance encoder blocks. For notational convenience, the output of $x_t$ processed through the U-Net encoder blocks is denoted as $x_{en}$, while the output of $m_r$ processed through the Room Guidance encoder blocks is denoted as $m_{en}$.
    $x_{en}$ and $m_{en}$ undergo linear transformations via three separate $1\times 1$ convolutional layers to generate the query $q$, key $k$, and value $v$, respectively. The output is computed using Eq. \ref{eq:eq02}, and the cross-attention mechanism is illustrated in Fig. \ref{fig:fig03}.
    \begin{equation}  
        \label{eq:eq02}
        \mathrm{CrossAttention}\left( q,k,v \right) =\mathrm{softmax} \left( {\frac{qk^T}{\sqrt{d_k}}} \right) v
    \end{equation}
    \item \textbf{\textit{Multi-scale Hierarchical Fusion and Semantic Attention}}: Inspired by RefineCatDiff \cite{liu2025refinecatdiff}, we construct the Room Guidance Encoder using a Conv-LayerNorm-ReLU architecture and extract multi-scale features from each layer, denoted as $r_e$, ensuring that its channel dimension matches that of the corresponding U-Net layer feature $x_e$.
    Subsequently, feature fusion is performed using Eq. \ref{eq:eq03}, where LN denotes LayerNorm, RF denotes Room Feature Fusion as in Fig. \ref{fig:fig02}.
    \begin{equation}  
        \label{eq:eq03}
        \begin{cases}
	    f=\mathrm{Concat}\left( x_e,r_e \right)\\
	    f^{\prime}=\mathrm{ReLU}\left[ \mathrm{LN}\left( \mathrm{Conv}\left( f \right) \right) \right]\\
	    \mathrm{RF}\left( x_e,r_e \right) =\mathrm{LN}\left( r_e\oplus f^{\prime} \right)\\
        \end{cases}
    \end{equation}
    To effectively inject high-level semantic features from the room guidance into the network, we design a Semantic Attention (SA) module that integrates both channel attention and spatial attention mechanisms.
    The SA module consists of parallel spatial and channel attention branches that jointly process $r_e$ and $x_e$, generating channel attention weights ($\alpha_{ch}$) and spatial attention weights ($\alpha_{sp}$), as shown in Fig. \ref{fig:fig04}. 
    Each branch performs element-wise multiplication between the input features and their respective attention weights. The refined features are then combined through element-wise addition, and the enhanced output is produced via a convolutional layer for further processing.
\end{itemize}

\subsubsection{Diffusion Learning}


In the diffusion process, the noisy data ${x_t}$ at step $t$ is calculated as follows:
\begin{equation}  
    \label{eq:eq1}
    {x_t=\sqrt{\hat{\beta}_t}\cdot x_0+\sqrt{\left( 1-\hat{\beta}_t \right)}\cdot z_t , z_t\sim \mathcal{N} \left( 0,\mathbf{I} \right)}
\end{equation}
Where ${\hat{\beta}_t=\prod\nolimits_{i=1}^t{\alpha _i}}$, ${\alpha _t=1-\beta _t}$ and ${t\in \left\{ 1,2,...,T \right\} }$, with the original map ${x_0}$ and a fixed variance schedule ${\beta _t}$, the noise ${x_t}$ can be sampled by Eq. \ref{eq:eq1}.

If the variance schedule ${\left( \beta _t \right) _{t=1}^{T}}$ makes ${\hat{\beta}_T\rightarrow 0}$, then ${x_T}$ approaches the standard Gaussian distribution, namely ${x_T\rightarrow \mathcal{N} \left( 0,\mathbf{I} \right)}$.
The training process is represented by reversing the forward diffusion process. By training a neural network model to predict the noise corresponding to each time step, the noise can then be removed one by one, gradually recovering the original data ${x_0}$.
This process is described as Eq. \ref{eq:eq5}, where $\theta$ represents the parameters of the neural network model, and the network takes noisy data ${x_t}$ and time step $t$ as input to predict the mean ${\mu _{\theta}\left( x_t,t,r \right)}$ and covariance ${\Sigma _{\theta}\left( x_t,t,r \right)}$.
Since our denoising process incorporates room guidance, the denoising model takes an additional input, denoted as $r$.
\begin{equation}  
    \label{eq:eq5}
    {p_{\theta}\left( x_{t-1} \middle| x_t \right) =\mathcal{N} \left( x_{t-1};\mu _{\theta}\left( x_t,t,r \right) ,\Sigma _{\theta}\left( x_t,t,r \right) \right) }
\end{equation}

Ho et al. \cite{ho2020denoising} proposed a simplified method, where the covariance ${\Sigma _{\theta}\left( x_t,t,r \right)}$ is fixed to a constant value and the mean ${\mu _{\theta}\left( x_t,t,r \right)}$ is rewritten as a function of noise, as follows:
\begin{equation}
    \label{eq:eq2}
    {\mu _{\theta}=\frac{1}{\sqrt{\alpha _t}}\cdot \left( x_t-\frac{1-\alpha _t}{\sqrt{1-\hat{\beta}_t}}\cdot z_{\theta}\left( x_t,t,r \right) \right) }
\end{equation}
With the simplified training objective, as follows:
\begin{equation}
    \label{eq:eq3}
    {\mathcal{L} _{vlb}=\mathbb{E} _{t\sim \left[ 1,T \right]}\mathbb{E} _{x_0\sim p\left( x_0 \right)}\mathbb{E} _{z_t\sim \mathcal{N} \left( 0,\mathbf{I} \right)}\left\| z_t-z_{\theta}\left( x_t,t,r \right) \right\| ^2}
\end{equation}

\begin{figure}[!t]
    \begin{center}
        \includegraphics[width=0.8\linewidth]{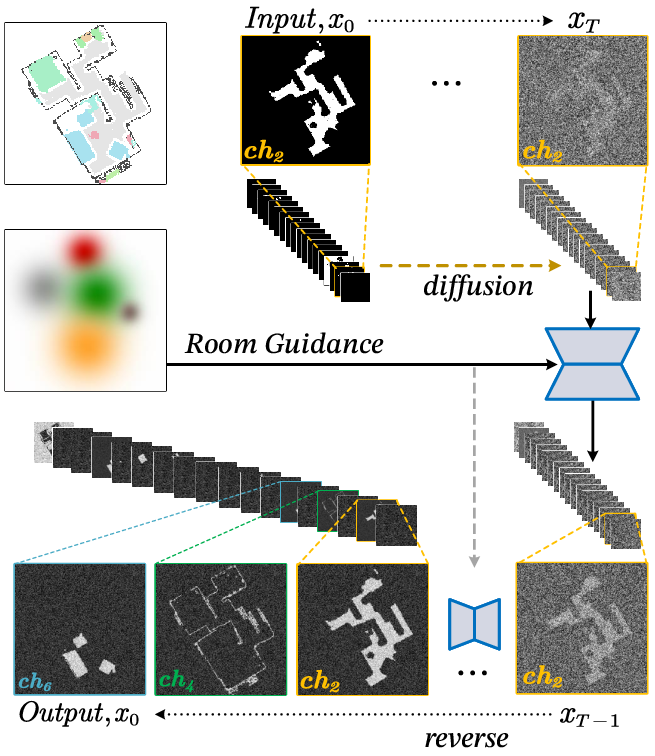}
    \end{center}
    \caption{
        Diffusion model learning process based on semantic maps, incorporating room guidance during denoising.
        }
    \label{fig:fig05}
\end{figure}
Where $\mathbb{E}$ is the expected value. As a result, the neural network model no longer needs to predict both the mean and covariance simultaneously, but only needs to estimate the noise ${z_{\theta}\left( x_t,t,r \right)}$ corresponding to ${x_t}$.

As shown in Fig. \ref{fig:fig05}, the input ${x_0}$ contains 18 binary channels (using the Gibson dataset as an example).
Following the standard training procedure of diffusion models for RGB images, the trained model can generate a denoised RGB image from noise. Similarly, when provided with 18-channel Gaussian noise as input, the model iteratively denoises the signal and eventually produces a semantic map.
Note that the model’s output is not a standard multi-channel binary semantic map by default. Thus, we apply an argmax operation as post-processing to discretize the predictions into valid semantic labels.

\begin{figure*}
    \begin{center}
        \includegraphics[width=\linewidth]{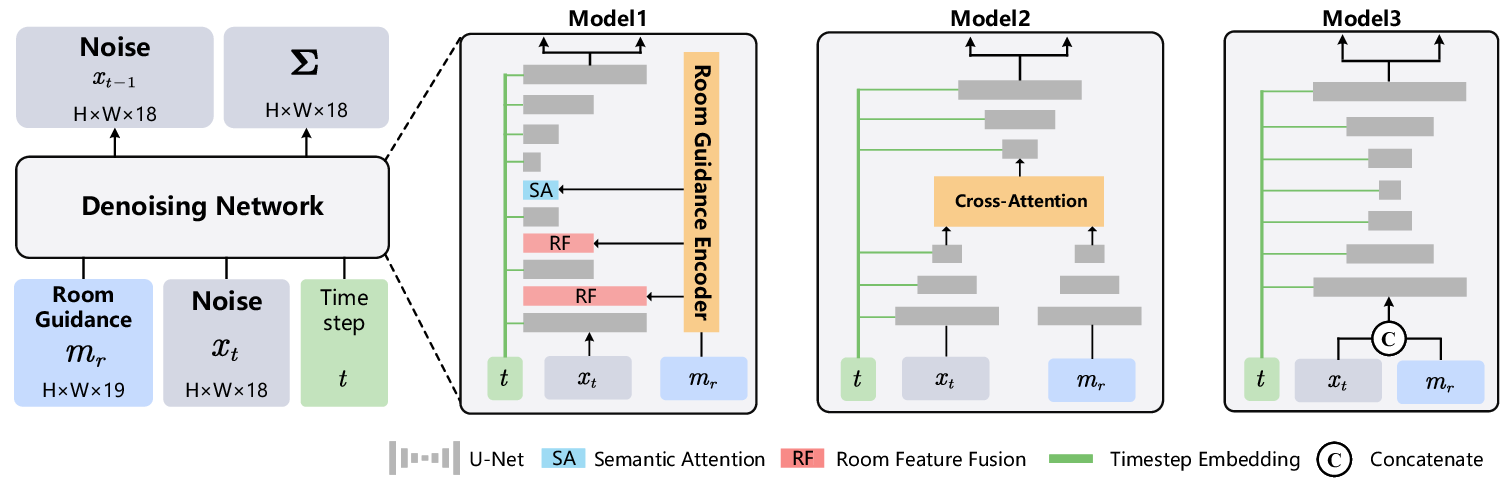}
    \end{center}
    \caption{
        Denoising model architecture. \textit{Left}: We train a conditional denoising network with three inputs, outputting noise and variance. \textit{Right}: Detailed model structure. We design and study three variants, incorporating room feature fusion, cross-attention mechanisms, and simple concatenation, respectively.
        }
    \label{fig:fig02}
\end{figure*}

\subsection{Unobserved regions generation Guided by LLM-Based Room Priors}
\label{sec:3-C}
\begin{figure*}
    \begin{center}
        \includegraphics[width=0.8\linewidth]{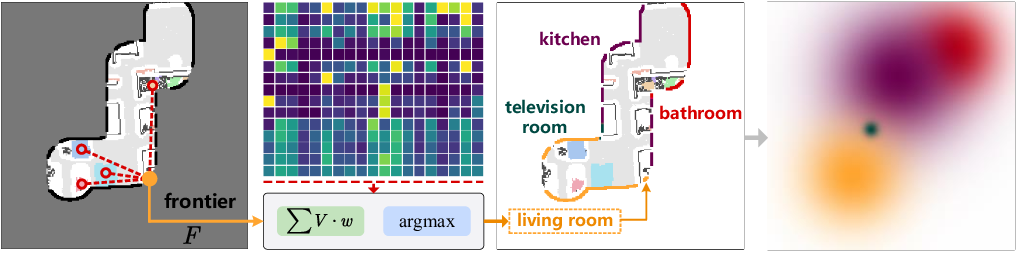}
    \end{center}
    \caption{
        Room guidance computation pipeline.
        }
    \label{fig:fig06}
\end{figure*}

We constructed a dataset in Section \ref{sec:dataset} to validate the map completion model. In this section, an example from this dataset is used for illustration. As shown in Fig. \ref{fig:fig06}, an object semantic map of the observed area was built during a navigation instance, with our goal being to generate the map content for unobserved regions. We first describe how to obtain LLM-based room priors for frontiers and finally explain how to generate the unknown area map using the pre-trained model from Section \ref{sec:3-B}.
\subsubsection{LLM-based Object-Room Priors}
To improve ObjectNav task efficiency, a key aspect is equipping the robot with semantic reasoning capability. Previous position prediction methods employed LLM-based object-to-object \cite{yu2023l3mvn}, room-to-room \cite{gao2023room}, and object-to-room \cite{sun2025enhancing} relationships to enhance navigation accuracy and generalization in unseen scenarios. However, these approaches applied commonsense knowledge to train the long-term goal prediction module, learning target relations individually. In contrast, we utilize LLM-based knowledge to guide map generation in unknown areas, employing object-room priors to direct a diffusion model in jointly learning multi-object relations—a method we demonstrate to be more reliable and efficient.

Similar to LROGNav \cite{sun2025enhancing} and LFG \cite{shah2023navigation}, we employ positive and negative prompts and utilize a chain-of-thought \cite{wei2022chain} approach to query the large language model (LLM) for determining the likelihood of each object belonging to specific room types.
This yields an $N \times M$ matrix $V$, where $N$ and $M$ represent the number of object and room categories, respectively. Here, $V_j\left( o \right)$ denotes the probability of object $o$ belonging to room type $j$. Next, our objective is to estimate the room categories associated with frontiers in the local map. Following PONI \cite{ramakrishnan2022poni}, we define a frontier as a non-obstacle region at the boundary between known and unknown areas.

As shown in the Fig. \ref{fig:fig06}, for frontier $F$, we select $n$ objects present in the map $\left[ o_1,o_2,...,o_n \right]$ ($n = 4$ in this case) and compute their corresponding weights $\left[ w_1,...,w_n \right]$ based on their distances to objects in the local map, where $\sum\nolimits_1^n{w_i}=1$.
We then compute the probability of frontier $F$ belonging to room category $j$ using Eq. \ref{eq:eq04}.
\begin{equation}
    \label{eq:eq04}
    R_j\left( F \right) =\sum\nolimits_{i=1}^n{V_j\left( o_i \right) \cdot w_i}
\end{equation}

Specifically, for each of the $n$ adjacent objects near $F$, we calculate its probability of belonging to room $j$ and multiply it by the corresponding weight. 
he final room assignment is determined by selecting the category with the highest probability, as formalized in Eq. \ref{eq:eq05}.
\begin{equation}
    \label{eq:eq05}
    R(F)=\underset{j\in 1,...,M}{\mathrm{argmax}} R_j(F)
\end{equation}

Next, we employ Eq. \ref{eq:eq01} to design a distribution map within each channel, where pixel values decay exponentially from the center outward, as illustrated in Fig. \ref{fig:fig06} .

\subsubsection{Map-Completion via Local Map-Conditioned Generation}

The observed local semantic map can be used as a condition to guide the diffusion model in generating the object distribution in the unknown regions, thereby enabling semantic reasoning of the target objects.
The local map established for the explored regions is denoted as ${x}$. After cropping and scaling preprocessing, the unknown area in ${x}$ constitute the mask ${m}$. 
The known grid is represented as ${x\odot \left( 1-m \right) }$, and the corresponding unknown region is denoted as ${x\odot m}$. We apply the RePaint \cite{lugmayr2022repaint} method, using Eq. \ref{eq:eq4} for the known region and Eq. \ref{eq:eq6} for the unknown region.
They are combined through the process represented by Eq. \ref{eq:eq7} and participate in the reverse process, ultimately generating a reasonable semantic content distribution in the unknown region of the local map.

\begin{equation}
    \label{eq:eq4}
    { x_{t-1}^{known}\sim \mathcal{N} \left( \sqrt{\hat{\beta}_t}\cdot x_0,\left( 1-\hat{\beta}_t \right) \cdot \mathbf{I} \right) }
\end{equation}
\begin{equation}
    \label{eq:eq6}
    { x_{t-1}^{unkown}\sim\mathcal{N} \left( \mu _{\theta}\left( x_t,t \right) ,\Sigma _{\theta}\left( x_t,t,r \right) \right) }
\end{equation}

\begin{equation}
    \label{eq:eq7}
    { x_{t-1}=\left( 1-m \right) \odot x_{t-1}^{known}+m\odot x_{t-1}^{unknown}}
\end{equation}

\begin{figure}[h]
    \begin{center}
        \includegraphics[width=\linewidth]{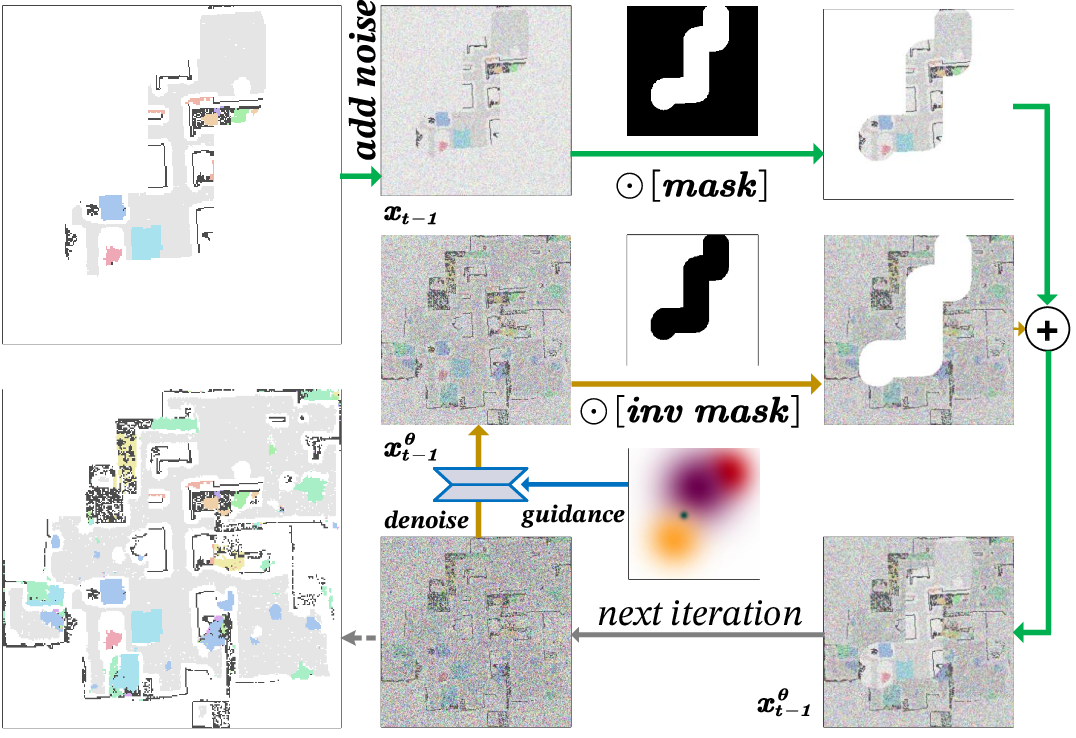}
    \end{center}
    \caption{
        Generating maps of unknown regions using the local map as a condition.
        }
    \label{fig:fig4}
\end{figure}
\begin{algorithm}
    \caption{Map completion}
    \label{alg:alg1}
    \begin{algorithmic}[1]
    \State \textbf{Input:} Total diffusion steps $T$, the Room Guidance $r$, trained denoising network $\psi _{\theta}\left( \cdot \right) $, local map $x_{local}$, map mask $m$.
    \State \textbf{Output:} The completed map $m^d$
    \State Generate sub-sequence of length $S$: $\left[ \tau _1,...,\tau _S \right] $
    \State Sample $x_{\tau _S}\sim \mathcal{N} \left( 0,\mathbf{I} \right) $
    \For {$t$ in $\left\{ \tau _S,\tau _{S-1},...,\tau _1 \right\} $}
        \State $\beta _t=\frac{10^{-4}\left( T-t \right) +2\times 10^{-2}\left( t-1 \right)}{T-1}$
        \State $\alpha _t=1-\beta _t$
        \State $\bar{\alpha}_t=\prod\nolimits_{n=1}^t{\alpha _n}$
        \State $\epsilon \sim \mathcal{N} \left( 0,\mathbf{I} \right) \,\,\mathrm{if}\ t>1, \mathrm{else}\ \epsilon =0$
        \State $x_{t-1}^{\mathrm{known}}=\sqrt{\bar{\alpha}_t}x_{local}+\left( 1-\bar{\alpha}_t \right) \epsilon$
        \State Estimate $x_0=\frac{x_t-\psi _{\theta}\left(x_t,t,r \right) \cdot \sqrt{1-\bar{\alpha}_t}}{\sqrt{\bar{\alpha}_t}}$
        \State Predict $\epsilon _t=\frac{x_t-\sqrt{\bar{\alpha}_t}x_0}{\sqrt{1-\bar{\alpha}_t}}$
        \State $x_{t-1}^{\mathrm{unknown}}=\sqrt{\bar{\alpha}_{t-1}}x_0+\sqrt{1-\bar{\alpha}_{t-1}}\epsilon _t $
        \State $x_{t-1}=m\odot x_{t-1}^{\mathrm{known}}+\left( 1-m \right) \odot x_{t-1}^{\mathrm{unknown}}$
    \EndFor
    \State \textbf{end for}
    \State \textbf{Return} $x_0$
    \end{algorithmic}
\end{algorithm}
As shown in Fig. \ref{fig:fig4}, ${\boldsymbol{x}_{\boldsymbol{t}-\boldsymbol{1}}}$ and ${\boldsymbol{x}_{\boldsymbol{t}-\boldsymbol{1}}^{\boldsymbol{\theta }}}$ represent the sampled result from the local map and the model trained in the previous section, respectively.

\subsection{Object Goal Navigation with DAR}

At each timestamp $t$ during navigation, the agent constructs a local semantic map ${m_t}$ based on sensor input. The semantic map contains multiple channels, with each channel composed of binary-valued pixels. The values in the unobserved regions are 0 across all channels.

\begin{figure*}[!h]
    \begin{center}
        \includegraphics[width=\linewidth]{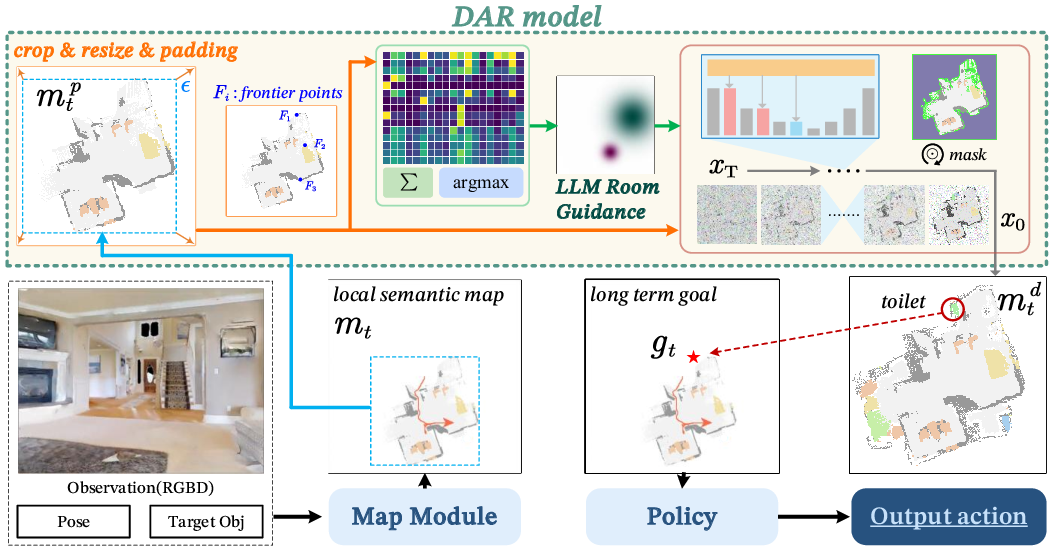}
    \end{center}
    \caption{
        During navigation, the agent builds a local semantic map ${m_t}$ from RGBD observations, which is processed through cropping, resizing, and padding to produce $m^p_t$.  The DAR model then generates a semantic map for the unknown areas in $m^p_t$, applying our proposed room guidance generation methods to manage noise in the denoising process. The long-term goal is set based on the generated map and the target.
        }
    \label{fig:fig7}
\end{figure*}

The trained diffusion model only allows for the prediction of map data at a resolution of ${256\times256}$. Therefore, we crop the minimum bounding box of the known region from the input ${m_t}$, as shown by the blue dashed box in Fig \ref{fig:fig7}. We then rescale the box region to a size of ${256/\epsilon}$ and pad it with zeros to ${256\times256}$, generating the processed map ${m_{t}^{p}}$.
The value of ${\epsilon}$  represents the range of unknown regions that the DAR model is used to predict. We set ${{\epsilon} = 120\%}$.

Next, using Eq. \ref{eq:eq04} and \ref{eq:eq05}, we estimate the room category for each frontier point based on pre-extracted object-room relations from the LLM, then generate room guidance using Eq. \ref{eq:eq01}.

Through the process described in Section \ref{sec:3-B}, the model progressively denoises the data and, under the guidance of room semantics, ultimately generates a semantic map ${m_{t}^{d}}$ within the unknown area of the ${m_{t}^{p}}$. 
In the ${m_{t}^{d}}$, non-zero values in the target channel indicate the predicted potential location of the target object by the DAR model. These coordinates are restored to the local map ${m_{t}}$ through inverse resizing and used as the long-term goal ${g_{t}}$.

It is worth noting that in the initial stage of navigation or when the sensors receive limited information, the map generated by the DAR model in the unknown regions may not contain the target object. In other words, the DAR model may not provide valuable information to the navigation system. In such cases, we set the long-term goal to the nearest frontier point.

Once the long-term goal is set, the agent just needs to move from its current location to the goal ${g_t}$. Following previous work \cite{chaplot2020object}, we use the Fast Marching Method \cite{sethian1996fast} to calculate the shortest path between the current location and the goal. The local policy then follows this path by taking deterministic actions to guide the agent.

Although diffusion models can effectively predict object distributions in unknown regions, their inherent nature results in slow DAR method inference. To address this limitation, we propose two parallel strategies to enhance map generation efficiency. First, inspired by DDIM \cite{song2020denoising}, we adopt an accelerated sampling method (Alg. \ref{alg:alg1}), downsampling the original $T$ diffusion steps to a subsequence of $S$ steps ($S < T$) to improve sampling speed. Second, instead of invoking DAR for long-term goal inference at every timestep, we employ a selective strategy: reasoning is triggered only when the agent reaches the vicinity of the long-term goal ($g_t$) or deviates significantly from it. Experimental results demonstrate the effectiveness of our approach—even with reduced inference frequency, the DAR model achieves leading performance improvements for object navigation systems.
\section{Experiment}

\subsection{Experimental setup}
\noindent\textbf{Dataset.} We evaluate our DAR on standard ObjectNav datasets: Gibson \cite{xia2018gibson} and Matterport3D (MP3D) \cite{chang2017matterport3d}. For Gibson and MP3D, we follow the setup of \cite{ramakrishnan2022poni} and \cite{zhang2024imagine}. Specifically, for Gibson, we use 25 training and 5 validation scenes from the Gibson tiny split. For MP3D, we use 56 training and 11 validation scenes, with 21 goal categories and 2,195 validation episodes. 

\noindent\textbf{Evaluation metrics.} For evaluating the navigation performance, we adopt three standard metrics following \cite{chaplot2020object} and \cite{ramakrishnan2022poni}: 1) SR: the ratio of success episodes. 2) SPL: the success rate weighted by the path length, which measures the efficiency of the path length. 3) DTS: the distance to the goal at the end of the episode.

\noindent\textbf{Implementation details.} 
For the self-supervised diffusion learning of DAR, we take room semantic maps from the Gibson and MP3D datasets, randomly rotate, scale, and crop them to $256\times256$ size for training samples. For training and evaluation, we use one NVIDIA RTX4090 GPU with 24GB memory.

\subsection{Comparisons with previous map completion methods.}
\label{sec:4-A}
\subsubsection{Map Completion Test DataSet}
\label{sec:dataset}
In this section, we construct a map completion test set that is independent of the navigation process and solely designed to evaluate the quality of algorithms in generating maps for unknown regions.
Taking Gibson as an example, we first obtain the semantic object map of a scene. Random translation and rotation are then applied as data augmentation, followed by cropping to a size of $256\times 256$. 
We then randomly select two locations in the navigable space of the complete map and compute the shortest path using the deterministic Fast Marching Method (FMM) \cite{sethian1996fast} path planner. For each location along this path, we sample an $S \times S$ (50 grids) square patch around it to represent the explored area. 
These patches are aggregated to form the local map $m^d$, while the corresponding map mask is also generated, where known and unknown regions are denoted by 1 and 0, respectively. The objective of the map completion algorithm is to generate the semantic map in unknown regions based on the map mask.
\begin{figure}
    \begin{center}
        \includegraphics[width=\linewidth]{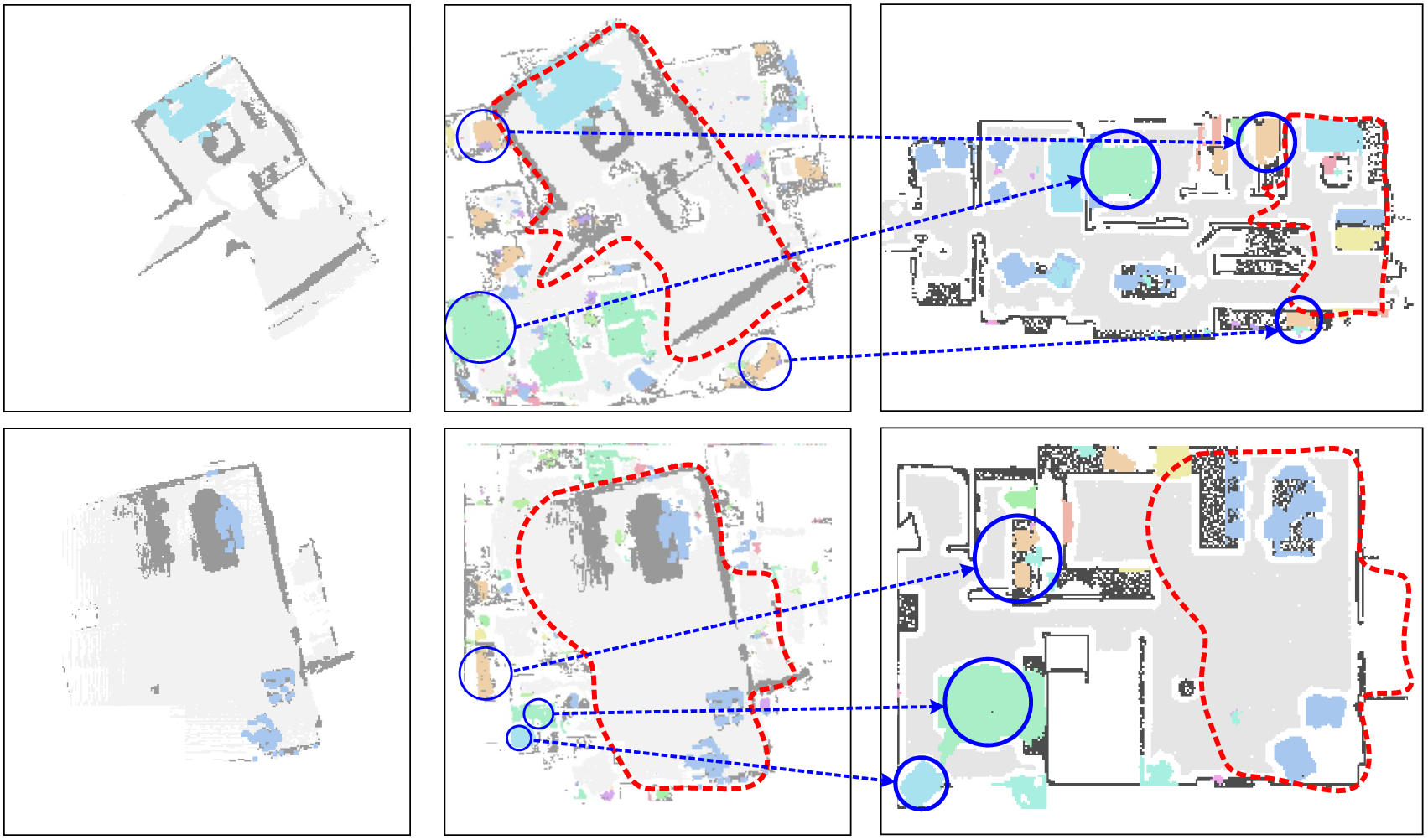}
    \end{center}
    \caption{
        Generated maps during navigation. As shown by the blue circles, the DAR model accurately predicts plausible positions of unobserved objects in unobserved areas.
        }
    \label{fig:fig12}
\end{figure}
To evaluate the quality of the generated maps, we employ two complementary metrics: the Intersection over Union (IoU), which measures per-pixel agreement across all semantic channels including both occupancy and object categories, and Category Recall, which quantifies the algorithm's ability to identify all relevant objects by computing the ratio of correctly predicted object categories to the total number of ground-truth categories. These metrics collectively assess both spatial accuracy and semantic recognition performance.

\subsubsection{Comparison results}
Previous works \cite{georgakis2021learning} train a UNet \cite{ronneberger2015u} to predict unobserved regions directly from a single semantic top-down map. Similar to our approach, SGM \cite{zhang2024imagine} leverages MAE \cite{he2022masked} to train and integrate LLMs' general knowledge for generating unobserved regions.


\begin{figure*}[!t]
    \begin{center}
        \includegraphics[width=\linewidth]{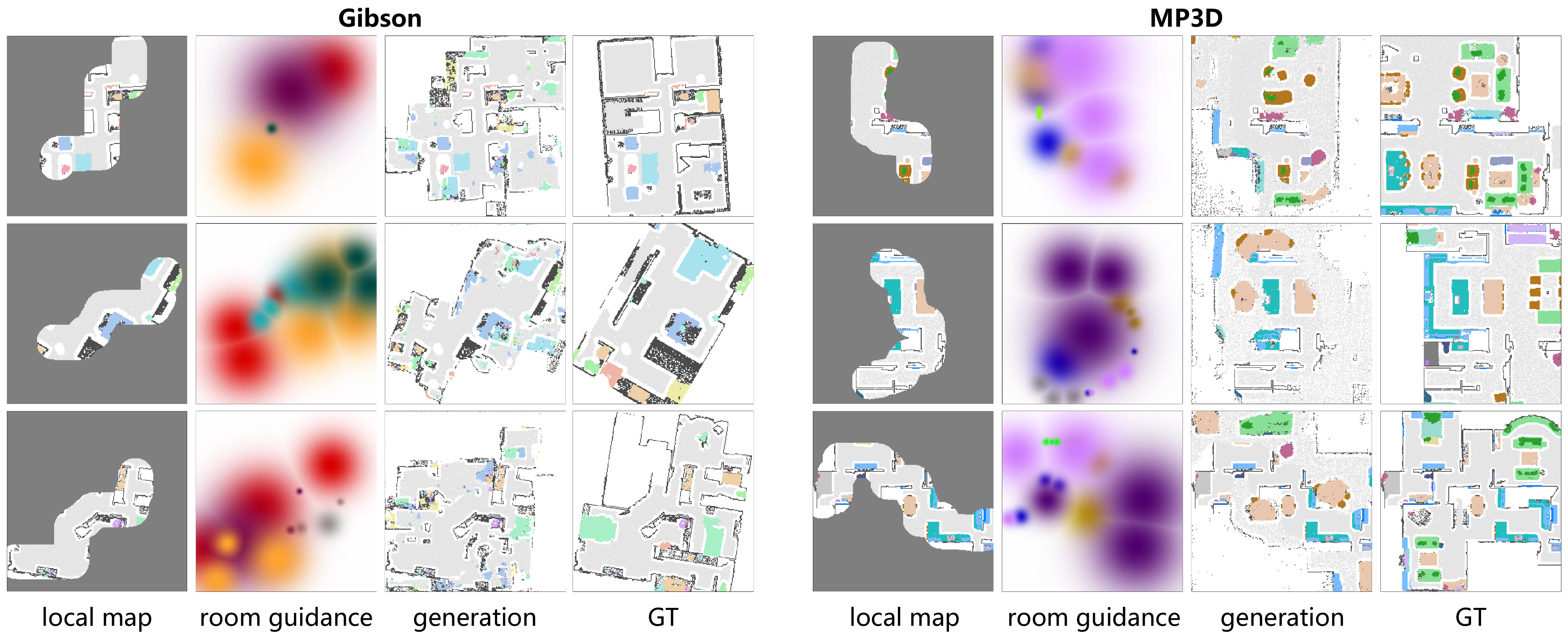}
    \end{center}
    \caption{
        Examples from the Map Completion Test Dataset, with corresponding room guidance and map completion results.
        }
    \label{fig:fig07}
\end{figure*}


We evaluated the map completion performance of different models on the Map Completion Test Dataset, which contains data more reflective of real-world navigation scenarios. While we consider MAE appropriate for model training, it is unsuitable for validating map completion performance because, in actual navigation, known regions are interconnected rather than widely scattered and disjointed.

As shown in Table. \ref{tab:tab1}, the Diffusion model significantly outperforms the ViT and CNN model. We attribute this to their inherent mechanisms. In the MAE+ViT training approach, the map is first divided into patches of a fixed size (e.g., 16 in SGM) to distinguish known and unknown regions. Due to computational and hardware constraints, the patch size cannot be set too small, leading to known-region patches containing some unknown pixels, which increases uncertainty in map generation. Moreover, the SGM method enhances completion performance by leveraging an LLM to provide general knowledge for reasoning about contextual relationships and predicting probable objects.

In this section, we also evaluated different implementations of Room-Guidance. Both the fixed $\sigma$ approach (in Eq. \ref{eq:eq01}) and the single-point method underperformed compared to the adaptive approach.
As shown in Fig. \ref{fig:fig07}, our DAR model successfully generates high-quality semantic maps in unknown regions. The object distribution aligns with room layouts and exhibits similarity to ground truth, demonstrating that the diffusion model can learn the distribution patterns of room objects and be effectively utilized for long-term goal reasoning in navigation.


\begin{table}
\centering
\caption{Performance comparison of different models on the Map Completion Test Dataset. (Highest scores are highlighted in \colorbox{first}{dark blue}, second-highest in \colorbox{second}{light blue}.)}
\label{tab:tab1}
\resizebox{\linewidth}{!}{%
\begin{tabular}{lcccc} 
\toprule
\multirow{2}{*}{Model}        & \multicolumn{2}{c}{Gibson} & \multicolumn{2}{c}{MP3D}  \\ 
\cmidrule{2-5}
                              & IoU$\uparrow$    & Recall$\uparrow$            & IoU$\uparrow$ & Recall$\uparrow$              \\ 
\midrule
MAE+ViT                       & 0.1670 & 0.2615            & 0.1424   & 0.3519                   \\
CNN                           & 0.1598 & 0.2542            & 0.1347   & 0.3465                   \\
SGM(MAE+ViT+LLM)              & 0.1833 & 0.2621            & 0.1683   & 0.3521                   \\ 
\hdashline[1pt/1pt]
Diff-M1+RG ($\sigma = 5$)                       & {\cellcolor[rgb]{0.667,0.863,0.878}}0.2607 & 0.2795            & {\cellcolor[rgb]{0.667,0.863,0.878}}0.2162   & {\cellcolor[rgb]{0.447,0.737,0.835}}0.3663                   \\
Diff-M2+RG ($\sigma = 5$)                       & 0.2365 & 0.2647            & 0.1879   & 0.3341                   \\
Diff-M3+RG ($\sigma = 5$)                       & 0.2181 & 0.2606            & 0.1840   & 0.3217                   \\ 
\hdashline[1pt/1pt]
Diff-M1+RG (point)     & 0.2543 & {\cellcolor[rgb]{0.447,0.737,0.835}}0.2892            & 0.2013  & 0.3562                   \\
Diff-M1+RG (self-adaptive)       & {\cellcolor[rgb]{0.447,0.737,0.835}}0.2728 & {\cellcolor[rgb]{0.667,0.863,0.878}}0.2808            & {\cellcolor[rgb]{0.447,0.737,0.835}}0.2276    & {\cellcolor[rgb]{0.667,0.863,0.878}}0.3603                     \\
\bottomrule
\end{tabular}
}
\end{table}

\subsection{Diffusion process analysis.}

\begin{figure}[htb]
    \begin{center}
        \includegraphics[width=\linewidth]{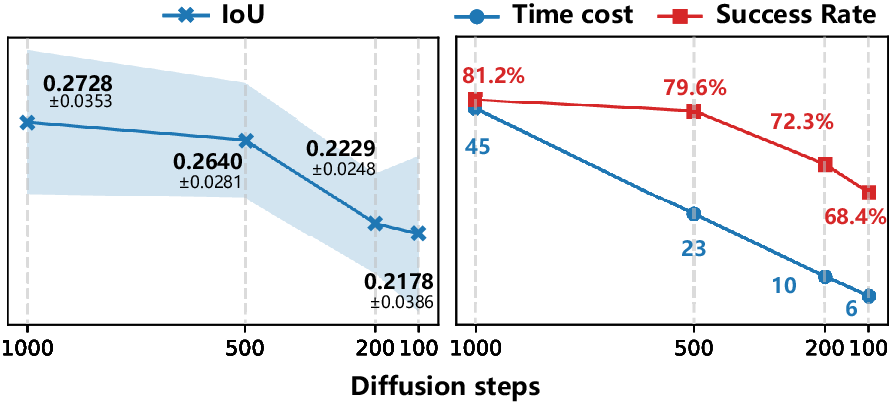}
    \end{center}
    \caption{
        Analysis of diffusion steps: Evaluation metrics include map generation IoU, time cost, and success rate on the Gibson validation navigation dataset.
        }
    \label{fig:fig11}
\end{figure}

In this section, we conduct a detailed investigation into how the diffusion steps affects model performance. Using the `Diff-M1+RG (self-adaptive)' configuration from Table. \ref{tab:tab1} and following the methodology in Alg. \ref{alg:alg1}, we vary the number of diffusion steps.
As shown in Fig. \ref{fig:fig11}, the generation time per sample exhibits an approximately linear relationship with diffusion steps.

Our experiments reveal a trade-off: reducing diffusion steps improves computational efficiency but leads to lower IoU in map generation and decreased navigation success rate (SR). Notably, when decreasing steps from 1000 to 500, we observe only a 2-3\% degradation in both IoU and SR metrics, while achieving a 49\% reduction in computation time. This suggests that 500 diffusion steps represent an optimal balance between computational efficiency and model performance.
In our implementation (Alg. \ref{alg:alg1}), we maintain $S=500$ and $T=1000$, the model is trained with 1000 steps but uses an accelerated sampling strategy that reduces inference steps to 500.

\subsection{Visualization of navigating with DAR.}
\begin{figure*}    
    \begin{center}
        \includegraphics[width=\linewidth]{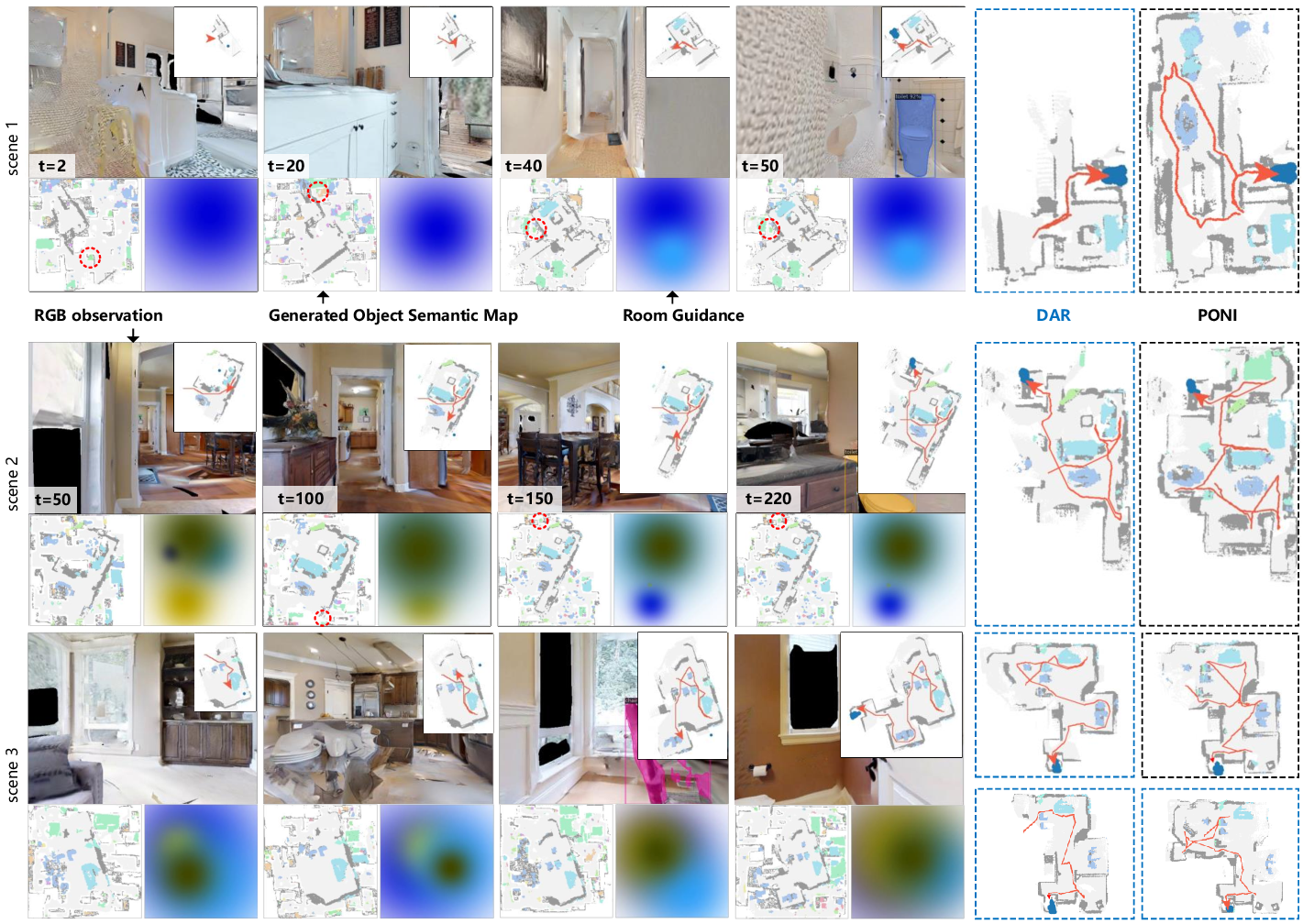}
    \end{center}
    \caption{
        Visualization of navigation using the proposed DAR method. The left four columns show the agent’s RGB view and local semantic map, including the trajectory and long-term goal (marked by a deep blue dot). It also displays the map generated by the DAR model. The right two columns show the path taken by the agent during navigation, demonstrating that the DAR method is more efficient than the PONI method.
        }
    \label{fig:fig9}
\end{figure*}

We follow the SGM configuration, and at the beginning of the navigation stage, the agent's collected environmental information is insufficient to support accurate visual reasoning, so we use the frontier-based exploration strategy \cite{luo2022stubborn}. As more scenes are observed, the DAR model successfully predicts the position of the target, even though it has not yet been observed.
During the navigation process, the DAR framework dynamically predicts object distributions in unexplored areas by leveraging the agent's accumulated environmental memory. This predictive capability enables DAR to generate long-term goals that effectively guide the exploration trajectory, as shown in Fig. \ref{fig:fig9}.

Notably, the diffusion model's inherent stochasticity introduces variability into DAR's semantic reasoning.
Consequently, even when operating in identical environments with the same initial conditions and target objectives, DAR exhibits non-deterministic behavior - each navigation query produces distinct exploratory paths and outcomes.
As shown in Fig. \ref{fig:fig9}, although the paths obtained are not completely identical, overall, DAR still significantly improves the efficiency of ObjectNav.
From Fig. \ref{fig:fig12}, it can also be seen that the DAR model not only successfully predicts the potential location of the target object during the navigation process, but also estimates the approximate locations of other objects and their context. There is still some deviation between the predicted target position and the true position by DAR, but we believe this is acceptable, as even humans cannot precisely locate unseen objects in an unseen scene.

These two visualizations (Fig. \ref{fig:fig9} and \ref{fig:fig12}) show that our DAR effectively expands limited local maps by generating unobserved regions, enabling the agent to infer the target’s location.

\subsection{Ablation study.}

\begin{table}
\centering
\caption{Performance of DAR model configurations on the Gibson and MP3D validation navigation datasets.}
\label{tab:tab2}
\resizebox{\linewidth}{!}{%
\begin{tabular}{lcccc} 
\toprule
\multirow{2}{*}{Model} & \multicolumn{2}{c}{Gibson}                       & \multicolumn{2}{c}{MP3D}                          \\ 
\cmidrule{2-5}
                       & \multicolumn{1}{c}{SR(\%)$\uparrow$} & \multicolumn{1}{c}{SPL(\%)$\uparrow$} & \multicolumn{1}{c}{SR(\%)$\uparrow$} & \multicolumn{1}{c}{SPL(\%)$\uparrow$}  \\ 
\midrule
Diff-M1 w/o RG         & 73.55                  & 40.7                    & 36.21                  & 13.1                     \\ 
\hdashline[1pt/1pt]
Diff-M1 w/o RF         & 79.64                  & 44.5                    & 39.03                  & 13.6                     \\
Diff-M1 w/o SA         & 80.18                  & {\cellcolor[rgb]{0.667,0.863,0.878}}44.6                    & 39.19                  & 13.9                     \\ 
\hdashline[1pt/1pt]
Diff-M3                & 78.51                  & 42.7                    & 37.25                  & 13.5                     \\
Diff-M2                & {\cellcolor[rgb]{0.667,0.863,0.878}}80.55                  & {\cellcolor[rgb]{0.667,0.863,0.878}}44.6                    & {\cellcolor[rgb]{0.667,0.863,0.878}}39.86                  & {\cellcolor[rgb]{0.667,0.863,0.878}}14.4                     \\
Diff-M1                & {\cellcolor[rgb]{0.447,0.737,0.835}}81.23                  & {\cellcolor[rgb]{0.447,0.737,0.835}}45.3                    & {\cellcolor[rgb]{0.447,0.737,0.835}}41.44                  & {\cellcolor[rgb]{0.447,0.737,0.835}}15.2                     \\
\bottomrule
\end{tabular}
}
\end{table}

In this section, we evaluate the performance of the three models proposed in Fig. \ref{fig:fig02} on the navigation dataset, measuring SR and SPL.
The results are presented in Table. \ref{tab:tab2}. Model 1 (M1) significantly outperforms the other two models. M3 employs the simplest room guidance fusion method but achieves the worst performance. Further ablation of M1 by removing the Semantic Attention (SA) module and Room Feature Fusion (RF) module leads to a performance drop of approximately 1\%\textasciitilde2\%.

As shown in the table, the use of room guidance leads to a significant performance improvement (accuracy metrics increase by 10\%\textasciitilde15\%). In the first row of the table ("w/o RG"), we disable all room guidance branches in Fig. \ref{fig:fig02}, retaining only the time step $t$ and noise $x_t$ as inputs.

As shown in Tables. \ref{tab:tab1} and \ref{tab:tab2}, if a map completion model achieves strong performance on our designed Map Completion Test Dataset, it also attains higher metrics on the navigation validation set. In conclusion, our experiments demonstrate the effectiveness of room guidance—leveraging LLM-inferred frontiers room categories to guide the diffusion model in generating object semantic maps for the corresponding regions significantly improves map completion accuracy and, consequently, Object Navigation success rates.

\subsection{Comparisons with the related works.}

\begin{table*}
\centering
\renewcommand{\arraystretch}{0.7}
\setlength{\extrarowheight}{0pt}
\addtolength{\extrarowheight}{\aboverulesep}
\addtolength{\extrarowheight}{\belowrulesep}
\setlength{\aboverulesep}{0pt}
\setlength{\belowrulesep}{0pt}
\caption{Performance comparison with other object navigation works.}
\label{tab:tab4}
\begin{tabular}{p{1cm};{1pt/1pt}>{\raggedright\arraybackslash}p{3cm};{1pt/1pt}>{\centering\arraybackslash}p{1.5cm}>{\centering\arraybackslash}p{1.5cm}>{\centering\arraybackslash}p{1.5cm};{1pt/1pt}>{\centering\arraybackslash}p{1.5cm}>{\centering\arraybackslash}p{1.5cm}>{\centering\arraybackslash}p{1.5cm}} 
\toprule
\multirow{2}{*}{ID} & \multirow{2}{*}{Methods} & \multicolumn{3}{c;{1pt/1pt}}{Gibson}                                                                                           & \multicolumn{3}{c}{MP3D}                                                                                                        \\ 
\cmidrule{3-8}
                    &                          & SR(\%)$\uparrow$                                       & SPL(\%)$\uparrow$                                      & DTS(m)$\downarrow$                                      & SR(\%)$\uparrow$                                       & SPL(\%)$\uparrow$                                      & DTS(m)$\downarrow$                                      \\ 
\midrule
\multirow{7}{*}{\uppercase\expandafter{\romannumeral1}}  & DD-PPO \cite{wijmans2019dd}                   & 15.0                                     & 10.7                                     & 3.24                                     & 8.0                                      & 1.8                                      & 6.94                                      \\
                    & Red-Rabbit \cite{ye2021auxiliary}              & \multicolumn{1}{c}{-}                    & \multicolumn{1}{c}{-}                    & \multicolumn{1}{c;{1pt/1pt}}{-}          & 34.6                                     & 7.9                                      & \multicolumn{1}{c}{-}                     \\
                    & THDA \cite{maksymets2021thda}                    & \multicolumn{1}{c}{-}                    & \multicolumn{1}{c}{-}                    & \multicolumn{1}{c;{1pt/1pt}}{-}          & 28.4                                     & 11.0                                     & 5.62                                      \\
                    & SSCNav \cite{liang2021sscnav}                   & \multicolumn{1}{c}{-}                                     & \multicolumn{1}{c}{-}                                     & \multicolumn{1}{c;{1pt/1pt}}{-}                                     & 27.1                                     & 11.2                                      & 5.71                                      \\
                    & Habitat-web \cite{ramrakhya2022habitat}              & \multicolumn{1}{c}{-}                    & \multicolumn{1}{c}{-}                    & \multicolumn{1}{c;{1pt/1pt}}{-}          & 35.4                                     & 10.2                                     & \multicolumn{1}{c}{-}                     \\
                    & EmbCLIP \cite{khandelwal2022simple}                  & 68.1                                     & 39.5                                     & 1.15                                     & 29.2                                     & 10.1                                     & 5.40                                      \\
                    & ENTL \cite{kotar2023entl}                     & \multicolumn{1}{c}{-}                    & \multicolumn{1}{c}{-}                    & \multicolumn{1}{c;{1pt/1pt}}{-}          & 17.0                                     & 5.0                                      & \multicolumn{1}{c}{-}                     \\ 
\hdashline[1pt/1pt]
\multirow{9}{*}{\uppercase\expandafter{\romannumeral2}}  & FBE \cite{yamauchi1997frontier}                      & 48.5                                     & 28.9                                     & 2.56                                     & 29.5                                     & 10.6                                     & 5.00                                      \\
                    & ANS \cite{chaplot2020neural}                      & 67.1                                     & 34.9                                     & 1.66                                     & 21.2                                     & 9.2                                      & 6.31                                      \\
                    & SemExp \cite{chaplot2020object}                   & 71.1                                     & 39.6                                     & 1.39                                     & 28.3                                     & 10.9                                     & 6.06                                      \\
                    & PONI \cite{ramakrishnan2022poni}                     & 73.6                                     & 41.0                                     & 1.25                                     & 27.8                                     & 12.1                                     & 5.63                                      \\
                    & L2M \cite{georgakis2021learning}                       & \multicolumn{1}{c}{-}                                     & \multicolumn{1}{c}{-}                                     & \multicolumn{1}{c;{1pt/1pt}}{-}                                     & 32.1                                     & 11.0                                    & 5.12                                      \\
                    & Stubborn \cite{luo2022stubborn}                 & \multicolumn{1}{c}{-}                    & \multicolumn{1}{c}{-}                    & \multicolumn{1}{c;{1pt/1pt}}{-}          & 31.2                                     & 13.5                                     & 5.01                                      \\
                    & L3MVN \cite{yu2023l3mvn}                    & 76.9                                     & 38.8                                     & {\cellcolor[rgb]{0.667,0.863,0.878}}1.01 & \multicolumn{1}{c}{-}                    & \multicolumn{1}{c}{-}                    & \multicolumn{1}{c}{-}                     \\
                    & SGM \cite{zhang2024imagine}                     & 78.0                                     & 44.0                                     & 1.11                                     & 37.7                                     & 14.7                                     & {\cellcolor[rgb]{0.667,0.863,0.878}}4.93  \\
                    & T-Diff \cite{yu2024trajectory}                   & {\cellcolor[rgb]{0.667,0.863,0.878}}79.6 & {\cellcolor[rgb]{0.667,0.863,0.878}}44.9 & {\cellcolor[rgb]{0.447,0.737,0.835}}1.00 & 39.6                                     & {\cellcolor[rgb]{0.447,0.737,0.835}}15.2 & 5.16                                      \\ 
\hdashline[1pt/1pt]
\multirow{2}{*}{\uppercase\expandafter{\romannumeral3}}  & DAR(0.5K)                & {\cellcolor[rgb]{0.667,0.863,0.878}}79.6 & 44.2                                     & 1.14                                     & {\cellcolor[rgb]{0.667,0.863,0.878}}40.6 & {\cellcolor[rgb]{0.667,0.863,0.878}}14.9 & 5.21                                      \\
                    & DAR(1K)                  & {\cellcolor[rgb]{0.447,0.737,0.835}}81.2 & {\cellcolor[rgb]{0.447,0.737,0.835}}45.3 & 1.12                                     & {\cellcolor[rgb]{0.447,0.737,0.835}}41.4 & {\cellcolor[rgb]{0.447,0.737,0.835}}15.2 & {\cellcolor[rgb]{0.447,0.737,0.835}}4.82  \\
\bottomrule
\end{tabular}
\end{table*} 

We categorize existing Object Navigation works, evaluated on either the Gibson or MP3D validation datasets, into two approaches:
1) End-to-End: Directly outputs actions, primarily using reinforcement learning (id \uppercase\expandafter{\romannumeral1}). 2) Modular-Based: Predicts a map-based long-term goal (id \uppercase\expandafter{\romannumeral2}). These are listed in Table \ref{tab:tab4}.
We briefly introduce the methods listed in the table as follows:

\setlist[itemize]{itemsep=0pt, leftmargin=*, label=\scriptsize$\bullet$}
\begin{itemize}
\item DD-PPO: Utilizes end-to-end reinforcement learning with distributed training.
\item Red-Rabbit: Enhances DD-PPO with auxiliary tasks for better sample efficiency and adaptability.
\item THDA: Introduces "Treasure Hunt Data Augmentation" to improve rewards and adaptability in new scenes.
\item SSCNav: an algorithm that explicitly models scene priors using a confidence-aware semantic scene completion module to complete the scene and guide the agent's navigation
\item Habitat-Web: Uses imitation learning with human demonstrations.
\item EmbCLIP: Integrates CLIP as a visual encoder into the model to predict the next action.
\item ENTL: A method for extracting long-sequence representations in embodied navigation, using vector-quantized predictions of future states conditioned on current states and actions.
\item FBE: Employs frontier-based exploration, creating a 2D occupancy map to find and approach targets.
\item ANS: Implements modular reinforcement learning to improve coverage, similar to FBE for target identification.
\item SemExp: A leading modular technique using reinforcement learning for effective long-term goal setting.
\item L2M: actively learns to generate semantic maps outside the field of view of the agent and leverages the uncertainty over the semantic classes in the unobserved areas to decide on long term goals.
\item Stubborn: Uses a modular approach, assigning objectives to map corners and adapting through rotation, integrating target detections over frames.
\item L3MVN: Leverages Large Language Models (LLMs) to incorporate commonsense reasoning for object searching, introducing two paradigms—zero-shot and feed-forward approaches—that use language to identify relevant frontiers from the semantic map as long-term goals.
\item SGM uses a self-supervised MAE method based on the ViT architecture and integrates pretrained RoBERTa to merge LLM predictions, generating a semantic map of unknown areas. It selects the location with the highest target confidence in the map as the long-term goal.
\item T-Diff: Learns the distribution of trajectory sequences conditioned on current observations and goal states.
\end{itemize}

Our proposed DAR model, based on a map-generation approach for long-term goal reasoning, outperforms both map-completion methods (e.g., SGM, SSCNav, and L2M) and location-prediction methods (e.g., PONI, SemExp, and T-Diff).
Notably, on the Gibson dataset, DAR surpasses the previous state-of-the-art (SOTA) supervised method, PONI, by 10.3\% in Success Rate (SR), 10.5\% in Success weighted by Path Length (SPL), and reduces Distance to Success (DTS) by 0.13 meters. Compared to the previous SOTA location-prediction method (T-Diff), DAR achieves nearly identical performance with 0.5K diffusion steps and further improves SR by 2\%\textasciitilde4\% with 1K steps.

\section{Conclusions}
\label{sec:limitation_conclusion}

In this work, we propose a modular-based self-supervised method for the object navigation task, named DAR. We demonstrate the effectiveness of applying diffusion models to lone-term goal reasoning.

We propose a Room Guidance Condition method integrated into the Denoising Diffusion Probabilistic Model (DDPM) framework for map generation and introduce a denoising model incorporating room guidance. Initially, we trained the DDPM on an indoor semantic map dataset, successfully generating high-quality and structurally plausible new map samples. This demonstrates that the trained DDPM effectively captures the statistical distribution patterns governing spatial relationships among indoor objects. Furthermore, we highlight that leveraging these patterns can significantly enhance the success rate and efficiency of agents in Object Navigation tasks.

Subsequently, we use Chain-of-Thought and positive-negative prompting techniques to query LLMs to obtain common sense knowledge, and incorporate it into the diffusion model's reasoning process through the proposed \textit{Room Guidance} method.

In summary, our DAR method explicitly leverages the structured knowledge from the Object-Room priors extracted from LLM and implicitly utilizes the unstructured knowledge from the pre-trained diffusion model. A series of experiments have shown that the DAR method achieves comparable performance to previous methods.
\section*{Acknowledge}
This work was supported by by the Young Scientists Fund of the National Natural Science Foundation of China (Grant No. 52105014) 
\bibliographystyle{cas-model2-names}

\bibliography{cas-refs}



\end{document}